%% file: main.tex
\documentclass[lettersize,journal]{IEEEtran}
\usepackage{amsmath,amsfonts}
\usepackage{algorithmic}
\usepackage{array}
\usepackage{booktabs}
\usepackage[caption=false,font=normalsize,labelfont=sf,textfont=sf]{subfig}
\usepackage{textcomp}
\usepackage{stfloats}
\usepackage{url}
\usepackage{verbatim}
\usepackage{graphicx}
\usepackage{caption}
\usepackage{xcolor}
\def\BibTeX{{\rm B\kern-.05em{\sc i\kern-.025em b}\kern-.08em
    T\kern-.1667em\lower.7ex\hbox{E}\kern-.125emX}}
\usepackage{balance}
\usepackage{makecell}
\usepackage{float} 

\newfloat{teaser}{tbp}{lop} 
\floatname{teaser}{Teaser}  

\usepackage{xurl}
\begin{document}
\title{View-Consistent 3D Scene Editing via Dual-Path Structural Correspondense and Semantic Continuity}

\author{Pufan Li, Bi'an Du,~\IEEEmembership{Student Member, IEEE}, Shenghe Zheng, Junyi Yao, Wei Hu,~\IEEEmembership{Senior~Member, ~IEEE}
\thanks{
 Pufan Li, Bi'an Du, Junyi Yao and Wei Hu are with Wangxuan Institute of Computer Technology, Peking University, No. 128, Zhongguancun North Street, Beijing, China (e-mail: lipufan@pku.edu.cn, pkudba@stu.pku.edu.cn, 2401112160@stu.pku.edu.cn,
 forhuwei@pku.edu.cn).

 Shenghe Zheng is with Department of Computer Science and Engineering, The Hong Kong University of Science and Technology, Clear Water Bay, Kowloon, Hong Kong SAR, China (e-mail: shenghez.zheng@gmail.com).
 
Corresponding author: Wei Hu (forhuwei@pku.edu.cn).
}
}

\markboth{Journal of \LaTeX\ Class Files,~Vol.~18, No.~9, September~2020}%
{How to Use the IEEEtran \LaTeX \ Templates}

\maketitle

  

\begin{abstract}
  \input{0_abstract}
\end{abstract}



\vspace{-0.15in}
\section{Introduction}

\input{1_introduction}

\vspace{-0.15in}
\section{Related Work}
\input{2_related_work}

\vspace{-0.15in}
\section{Method}
\input{3_method}

\vspace{-0.1in}
\section{Experiments}
\input{4_experiments}

\vspace{-0.1in}
\section{Conclusion}
\input{5_conclusion}
\bibliographystyle{ieeetr}
\vspace{-0.2in}
\bibliography{reference.bib}  

\end{document}

%% file: 0_abstract.tex
Text-driven 3D scene editing has recently attracted increasing attention. Most existing methods follow a render-edit-optimize pipeline, where multi-view images are rendered from a 3D scene, edited with 2D image editors, and then used to optimize the underlying 3D representation. However, cross-view inconsistency remains a major bottleneck. Although recent methods introduce geometric cues, cross-view interactions, or video priors to mitigate this issue, they still largely rely on inference-time synchronization and thus remain limited in robustness and generalization.
In this work, we recast multi-view consistent 3D editing from a distributional perspective: 3D scene editing essentially requires a joint distribution modeling across viewpoints.
Based on this insight, we propose a view-consistent 3D editing framework that explicitly introduces cross-view dependencies into the editing process. 
Furthermore, motivated by the observation that structural correspondence and semantic continuity rely on different cross-view cues, we introduce a dual-path consistency mechanism consisting of projection-guided structural guidance and patch-level semantic propagation for effective cross-view editing. 
Further, we construct a paired multi-view editing dataset that provides reliable supervision for learning cross-view consistency in edited scenes. Extensive experiments demonstrate that  our method achieves superior editing performance with precise and consistent views for complex scenes.


%% file: 1_introduction.tex
Text-driven 3D scene editing aims to modify a 3D scene according to natural-language instructions \cite{haque2023instruct, liu2021editing, wu2024gaussctrl, luo2025trame, lee2025editsplat, chen2024dge, chen2026fast}, enabling intuitive and controllable content creation. This enabled various applications such as virtual production, digital asset editing, and immersive environment design.

\begin{figure}[t]
  \centering
    \includegraphics[width=0.48\textwidth]{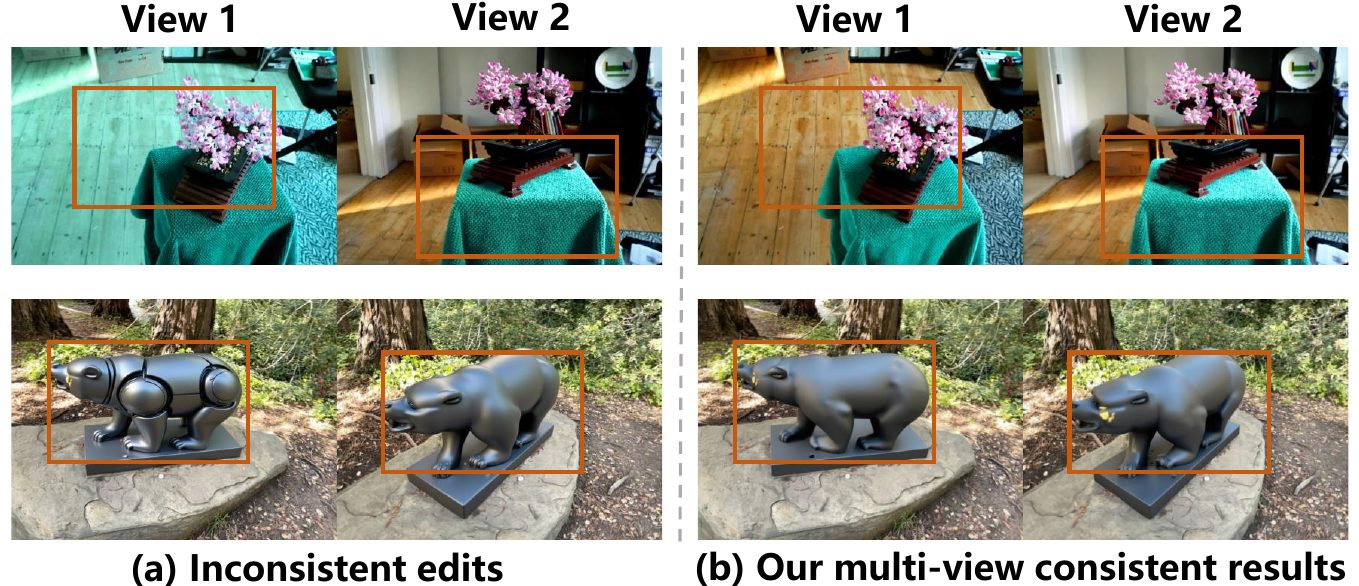}
  \caption {(a) Cross-view discrepancies in per-view edits, highlighted in orange boxes. (b) Our multi-view consistent results, which preserve structural correspondence and semantic continuity across viewpoints.}
  \label{fig:pic1}
  \vspace{-0.2in}
\end{figure}

Most recent 3D editing methods adopt a common pipeline: they first render a set of images from the 3D scene, edit these views with a pre-trained 2D diffusion model, and then optimize the underlying 3D representation using the edited results. 
Earlier approaches mainly differ in how the edited views are incorporated back into the 3D scene, such as through iterative dataset updates \cite{haque2023instruct, chen2024gaussianeditor} or diffusion-based guidance and optimization \cite{liu2021editing, hong2025perturb}. 
Despite these differences, they still rely primarily on per-view 2D priors, without explicitly modeling consistency across viewpoints.
In practice, however, such per-view 2D priors do not naturally enforce multi-view consistency.
As a result, the edited results may exhibit mismatched geometry, appearance, or semantic details. 
As shown in Fig.~\ref{fig:pic1} (a), such inconsistencies often appear as cross-view discrepancies in local structure and appearance.

To address this, previous methods leverage cross-view interaction to improve consistency across viewpoints.
GaussCtrl \cite{wu2024gaussctrl} introduces geometric conditions through depth-conditioned editing and latent alignment to improve geometry and appearance consistency. 
EditSplat \cite{lee2025editsplat} fuses neighboring view information through projection, blending, and attention mechanisms, and further improves optimization with attention-guided trimming.
DGE \cite{chen2024dge} adopts spatio-temporal attention with geometric constraints to obtain view-consistent edited sequences. 
ViP3DE \cite{chen2026fast} leverages video priors to improve multi-view coherence through cross-frame consistency cues on top of the image editing pipeline. 
Nevertheless, these methods reveal a fundamental issue: current 3D editing pipelines are still built upon single-image editing models with independent distribution modeling, while the actual target is multi-view coherent editing, requiring {\it joint distribution modeling}.

To this end, we propose a cross-view consistency framework for text-driven 3D scene editing, which explicitly models the joint distribution of multiple views, 
aiming to enhance multi-view consistency. Instead of imposing consistency only through inference-time synchronization, our framework learns it as an intrinsic property of multi-view editing, enabling a strong single-image editor to operate reliably across views.
Our key observation is that effective cross-view consistency depends on two distinct yet coupled requirements: {1)} \textit{structural correspondence}, which preserves spatial alignment across viewpoints; and {2)} \textit{semantic continuity}, which maintains stable and coherent edited content across views. These two requirements rely on different forms of cross-view cues. Specifically, the structural consistency requires \textit{geometry-aware guidance} to preserve spatial correspondence, whereas the semantic consistency requires \textit{cross-view semantic context} to maintain stable edited semantics across viewpoints.
Motivated by this observation, we design a dual-path consistency mechanism. Concretely, we introduce two modules: {1)} a \textit{projection-guided structural guidance} path to enforce cross-view correspondence in structure, which encodes projected structural cues and adds the resulting guidance features to the corresponding backbone activations; and {2)} a \textit{patch-level semantic propagation} path to enforce cross-view continuity in semantics, which propagates semantic context from the previous edited result via reference-guided attention. Altogether, the two paths enable coherent multi-view editing while preserving the strong editing capability of the backbone diffusion model. 
As shown in Fig.~\ref{fig:pic1} (b), the proposed framework produces coherent editing results across viewpoints by jointly preserving structural correspondence and semantic continuity.

To support the learning of cross-view consistency, a key challenge lies in the lack of suitable datasets for supervision. 
We observe that large image editing models can already produce strong pairwise consistency when adjacent views are concatenated and jointly edited, which motivates us to construct \textit{Cross-View Consistency Editing Dataset} (\textit{CVC-Edit}), a paired multi-view editing dataset for cross-view propagation learning. 
To improve the reliability of this dataset, we further apply consistency-aware filtering based on global and local features. 
The resulting dataset provides reliable supervision for learning cross-view consistency in edited scenes, and offers a useful resource for multi-view scene editing research. 
Extensive experiments verify the effectiveness of our framework in producing precise and view-consistent editing results for complex scenes.

Our main contributions can be summarized as follows:

\begin{itemize}


\item We propose a cross-view consistency framework for 3D editing by formulating multi-view editing as the modeling of a joint cross-view distribution, rather than independent single-image editing, aiming to alleviate multi-view inconsistency.
  
\item We introduce a dual-path consistency mechanism for cross-view editing, consisting of projection-guided structural guidance and patch-level semantic propagation, which explicitly model the structural correspondence and semantic continuity across viewpoints.

\item We construct a consistency-aware paired multi-view editing dataset \textit{CVC-Edit}, which provides reliable supervision for learning cross-view consistency in edited scenes. Experimental results show that the proposed method achieves superior editing performance with multi-view consistency. 

\end{itemize}


%% file: 2_related_work.tex
\subsection{Diffusion Models}
\vspace{-0.05in}

Diffusion models have become a dominant paradigm for visual generation due to their strong image fidelity, controllability, and scalability \cite{ho2020denoising, song2020denoising, rombach2022high, chen2026tuning, luo2026diffw, V-Bridge, chen2022deep}. 
They have established a standard foundation for text-to-image and multimodal generation \cite{zhang2025multi, hou2025clip}, while also providing flexible conditioning mechanisms and strong semantic priors for a wide range of downstream tasks \cite{zhang2023adding, ye2023ip}. 
Recent advances have further improved diffusion-based modeling from both architectural and training perspectives: 
Diffusion Transformers replace convolutional U-Net backbones with transformer-based denoisers, showing strong scalability and generation quality \cite{peebles2023scalable, vaswani2017attention, zheng2025free, zheng2024dclp}, 
while Flow Matching formulates generation as learning continuous transport dynamics through vector field regression \cite{albergo2023building, lipmanflow}. 
Beyond 2D synthesis, diffusion models have also been widely extended to 3D generation, either by exploiting pretrained 2D diffusion priors to guide the generation of 3D shapes and radiance fields \cite{li2025geometry}, or by directly learning diffusion models in 3D space for native 3D generation \cite{luo2021diffusion, vahdat2022lion, nichol2022point, du2026part, du2024multi, du2024generative}.
These developments establish diffusion models as a powerful backbone for conditional generation and 3D content creation.

\vspace{-0.15in}
\subsection{Image Editing Foundation Models}
\vspace{-0.05in}

With the rapid progress of diffusion-based generation, image editing models have evolved from task-specific systems to increasingly general-purpose foundation models. Early diffusion-based editing methods relied on inversion, prompt manipulation, or attention control to modify image content under textual or structural guidance \cite{mokady2023null, tumanyan2023plug}, while other works explored more specialized editing settings such as face editing, exemplar-guided manipulation, and external attention control \cite{hou2022textface, ak2022learning, gao2022external}. InstructPix2Pix \cite{brooks2023instructpix2pix} further established a practical paradigm for free-form instruction-guided editing by training on synthetic editing triplets, greatly simplifying the editing pipeline and making language-driven editing more scalable. Building on this paradigm, recent image editing foundation models \cite{labs2025flux, wu2025qwen} have substantially improved semantic precision, editing flexibility, and instruction-following capability, enabling more complex, compositional, and scene-level modifications. 

\vspace{-0.15in}
\subsection{Text-Driven 3D Scene Editing}
\vspace{-0.05in}

Text-driven 3D scene editing aims to modify a reconstructed 3D scene according to natural language instructions \cite{wu2024gaussctrl, lee2025editsplat, chen2024dge, chen2026fast, yu2024morphnerf}. 
Existing methods mainly follow two routes. 
One line directly optimizes the 3D representation under the guidance of pretrained 2D models, such as CLIP- or diffusion-based supervision \cite{liu2021editing, hong2025perturb}. 
These methods avoid explicit image-space editing, but often suffer from limited fidelity or unstable optimization due to the indirect guidance signal.

The other line adopts a render-edit-optimize pipeline, where a set of rendered views is first edited by a 2D image editor and then used to update the 3D representation \cite{haque2023instruct, wu2024gaussctrl, lee2025editsplat, chen2024dge, chen2026fast,chen2024gaussianeditor, du2024auggs}. 
This paradigm has become a practical solution for text-driven 3D editing because it transfers strong 2D editing priors to 3D tasks without requiring paired 3D supervision. 
With the emergence of 3D Gaussian Splatting, recent studies have further improved editing efficiency through its explicit representation and fast rendering capability \cite{kerbl20233d, wu2024gaussctrl, lee2025editsplat, chen2024dge}.
To improve cross-view consistency, existing methods have introduced depth-conditioned editing \cite{wu2024gaussctrl}, projection-based fusion \cite{lee2025editsplat}, spatio-temporal attention \cite{chen2024dge}, and video priors \cite{chen2026fast}. 
Despite these advances, maintaining robust cross-view consistency remains a central challenge in text-driven 3D scene editing. Although these methods introduce various strategies to enhance consistency at inference time, the underlying 2D editors are still primarily trained for single-image manipulation rather than cross-view coherent editing, which limits their robustness and generalization.

%% file: 3_method.tex
\vspace{-0.1in}
\begin{figure*}[htbp]
  \centering
  \includegraphics[width=\textwidth]{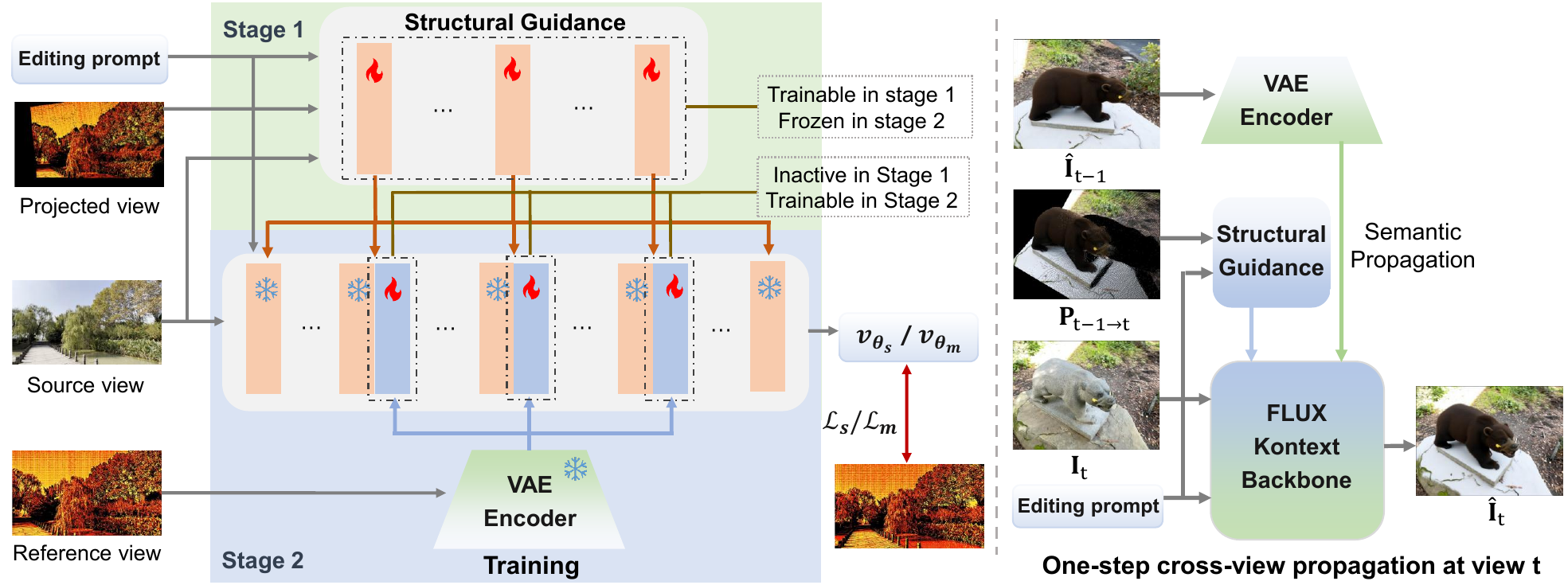}
  \vspace{-0.2in}
\caption{
Overview of the proposed framework. 
Left: two-stage training under a unified architecture, where Stage 1 trains the projection-guided structural guidance path ($\theta_s$) and Stage 2 freezes it and trains the patch-level semantic propagation path ($\theta_m$). 
Right: inference at view $t$, where $\mathbf{P}_{t-1\rightarrow t}$ and $\hat{\mathbf{I}}_{t-1}$ are fed into the structural guidance and semantic propagation paths respectively, jointly guiding the editing of the current view to produce $\hat{\mathbf{I}}_t$ consistent with the reference cues.
}
  \label{fig:framework2}
  \vspace{-0.2in}
\end{figure*}
Given an original 3D Gaussian Splatting (3DGS)  $\mathcal{G}$ as 3D representation, and a text prompt $c$, our goal is to transform $\mathcal{G}$ into an edited 3DGS $\mathcal{G}'$ that better matches the target content specified by the prompt.

We will first introduce the preliminary in Sec.~\ref{preliminary}. Then we elaborate on the proposed view-consistent editing framework in Sec.~\ref{frame}, which includes two key components: projection-guided structural guidance in Sec. \ref{structure}, and patch-level semantic propagation in Sec. \ref{semantic}. Sec. \ref{data} describes the data construction process that provides supervision for these modules, and Sec. \ref{training} presents the training and inference pipelines. 

\vspace{-0.2in}
\subsection{Preliminary}
\vspace{-0.05in}
\label{preliminary}
\subsubsection{3D Gaussian Splatting}

3D Gaussian Splatting (3DGS) \cite{kerbl20233d} is an efficient 3D scene representation and rendering technique that has recently achieved strong performance in both quality and speed. It represents a scene as a set of anisotropic 3D Gaussians and enables fast differentiable rendering through splatting-based rasterization. The $i$-th Gaussian is parameterized by its center position $\mathbf{x}_i \in \mathbb{R}^3$, scaling factor $\mathbf{s}_i \in \mathbb{R}^3$, rotation quaternion $\mathbf{q}_i \in \mathbb{R}^4$, opacity $\alpha \in \mathbb{R}$, and color attribute $\mathbf{c}_i \in \mathbb{R}^3$. The scaling and rotation jointly determine the Gaussian covariance. Altogether, we denote the parameters of the $i$-th Gaussian by:
\[
\Theta_i = \{\mathbf{x}_i, \mathbf{s}_i, \mathbf{q}_i, \alpha_i, \mathbf{c}_i\},
\]
and use $\Theta$ to represent the full set of Gaussians in the scene.

Given a camera view, the color of each pixel $\mathbf{p}$ is obtained by alpha compositing the projected Gaussians in depth order:
\begin{equation}
\mathbf{C}(\mathbf{p})=\sum_{i\in\mathcal{N}} \mathbf{c}_i \alpha_i \prod_{j=1}^{i-1}(1-\alpha_j),
\label{eq:gs_render}
\end{equation}
where $\mathcal{N}$ denotes the set of Gaussians contributing to pixel $\mathbf{p}$.

\subsubsection{Direct Gaussian Editor} 
\label{DirectEdit}

A direct Gaussian editor follows a render-edit-optimize pipeline. Given an original 3DGS scene representation $\mathcal{G}$ and a set of camera parameters $\{\pi_i\}_{i=1}^{N}$, the scene is first rendered into multi-view images:
\begin{equation}
\mathbf{I}_i = Rend(\mathcal{G}, \pi_i), \quad i=1,\dots,N,
\end{equation}
where $Rend(\cdot)$ denotes the differentiable renderer of 3DGS. A 2D image editor $\mathcal{E}(\cdot,\cdot)$ is then applied to each rendered view independently, producing edited images:
\begin{equation}
\hat{\mathbf{I}}_i = \mathcal{E}(\mathbf{I}_i,c),
\end{equation}
where $c$ is the editing prompt. Finally, the edited views are used as supervisions to optimize the Gaussian parameters:
\begin{equation}
\mathcal{G}' = \arg\min_{\mathcal{G}} \sum_{i=1}^{N} \mathcal{L}\big(Rend(\mathcal{G}, \pi_i), \hat{\mathbf{I}}_i\big),
\label{eq:optimizegs}
\end{equation}
where $\mathcal{L}$ denotes the reconstruction objective. Although this formulation provides a practical way to transfer strong 2D editing priors to 3DGS editing, independently edited views often introduce inconsistent supervision across viewpoints.

\vspace{-0.2in}

\subsection{The Proposed Cross-View Consistency Framework}
\vspace{-0.05in}
\label{frame}
The formulation in Sec.~\ref{DirectEdit} implicitly treats multi-view editing as independent single-image editing.
Essentially, it approximates the target distribution as:
\begin{equation}
p(\{\hat{\mathbf{I}}_i\}_{i=1}^{N} \mid \{\mathbf{I}_i\}_{i=1}^{N}, c)
\approx
\prod_{i=1}^{N} p(\hat{\mathbf{I}}_i \mid \mathbf{I}_i, c).
\end{equation}
However, the goal of 3D scene editing is not to obtain a set of individually plausible edited images, but to generate a set of \emph{jointly coherent} views corresponding to the same edited scene. This exposes a mismatch between the native prior of image editing models which is learned under an independent single-image editing distribution and the joint cross-view coherent distribution required for multi-view 3D scene editing.

To reduce this mismatch, we formulate multi-view editing as a consistency-aware process with explicit cross-view dependencies. Since the edited views correspond to different observations of the same edited scene, they should be modeled by a joint conditional distribution rather than as independent outputs. By the chain rule of conditional probability, this joint distribution can be exactly factorized as:
\begin{equation}
\begin{aligned}
p(\{\hat{\mathbf{I}}_i\}_{i=1}^{N} \mid \{\mathbf{I}_i\}_{i=1}^{N}, c)
&=
p(\hat{\mathbf{I}}_1 \mid \{\mathbf{I}_i\}_{i=1}^{N}, c) \\
&\quad \cdot
\prod_{i=2}^{N}
p(\hat{\mathbf{I}}_i \mid \hat{\mathbf{I}}_{<i}, \{\mathbf{I}_i\}_{i=1}^{N}, c),
\end{aligned}
\vspace{-0.1in}
\end{equation}
where $\hat{\mathbf{I}}_{<i}=\{\hat{\mathbf{I}}_1,\dots,\hat{\mathbf{I}}_{i-1}\}$ denotes the edited views preceding the $i$-th view. This factorization is exact and follows directly from the definition of conditional probability. However, directly modeling such full conditional dependencies is considerably less scalable, as it requires maintaining and interacting over high-dimensional intermediate representations from multiple views simultaneously. Instead, we adopt a tractable neighboring-view approximation in which edited information is progressively propagated across adjacent viewpoints:
\begin{equation}
p(\{\hat{\mathbf{I}}_i\}_{i=1}^{N} \mid \{\mathbf{I}_i\}_{i=1}^{N}, c)
\approx
p(\hat{\mathbf{I}}_1 \mid \mathbf{I}_1, c)
\prod_{i=2}^{N}
p(\hat{\mathbf{I}}_i \mid \mathbf{I}_i, \hat{\mathbf{I}}_{i-1}, c).
\end{equation}

Here, the full history $\hat{\mathbf{I}}_{<i}$ is approximated by the most recent edited view $\hat{\mathbf{I}}_{i-1}$, and the dependence on all input views is reduced to $\mathbf{I}_i$. This first-order approximation is motivated by two factors: adjacent views share the largest overlap and the most reliable local correspondences, and the previous edited view already carries information propagated from earlier views. Under this formulation, each target view is conditioned on the current view, the text prompt, and the propagated neighboring-view state.

Moreover, cross-view consistency involves two complementary aspects. The first is \emph{structural consistency}, which preserves spatial correspondence across viewpoints so that local regions, boundaries, and scene layouts remain aligned under view changes. The second is \emph{semantic consistency}, which preserves the edited semantic state so that object identity, attributes, materials, and overall appearance remain stable across viewpoints. Since these two aspects characterize different facets of cross-view editing, we model them separately. This leads to our dual-path consistency framework, consisting of a projection-guided structural guidance path and a patch-level semantic propagation path.
An overview of the proposed framework is shown in Fig.~\ref{fig:framework2}. 
The following two subsections describe these two components in detail.

\vspace{-0.15in}
\subsection{Projection-Guided Structural Guidance}
\vspace{-0.05in}
\label{structure}

To explicitly enforce structural consistency across viewpoints, we introduce a projection-guided structural guidance path that transfers geometry-aware editing cues between neighboring views. 
The key idea is to reproject the previously edited view into the current viewpoint using estimated scene depth and relative camera transformation, so as to provide a structure-aware hypothesis of how the edited content should spatially correspond under the new view, thereby providing explicit structural guidance for the editing process. 

Formally, let $\mathbf{I}_{i-1}$ and $\mathbf{I}_i$ denote two neighboring rendered views from the same 3DGS scene, and let $\hat{\mathbf{I}}_{i-1}$ be the edited result of $\mathbf{I}_{i-1}$ under a text prompt $c$. We estimate the depth map $\mathbf{D}_{i-1}$ for $\mathbf{I}_{i-1}$ using Depth-Anything-3 \cite{lin2025depth}, and obtain the relative camera transformation $\mathbf{T}_{i-1 \rightarrow i}$ between the two views. Based on these, we warp the edited image $\hat{\mathbf{I}}_{i-1}$ into the target view to obtain the projected structural cue:
\begin{equation}
\mathbf{P}_{i-1 \rightarrow i} = \mathcal{W}(\hat{\mathbf{I}}_{i-1}, \mathbf{D}_{i-1}, \mathbf{T}_{i-1 \rightarrow i}),
\end{equation}
where $\mathcal{W}(\cdot)$ denotes a  warping operation.

The projected image $\mathbf{P}_{i-1 \rightarrow i}$ provides an explicit hypothesis of how the edited structure should appear in $\mathbf{I}_i$, and thus serves as a geometry-aware control signal for the editing model. Compared with purely inference-time synchronization strategies, this projection introduces physically grounded cross-view correspondence based on scene geometry. However, due to depth estimation errors, occlusion, and view-dependent appearance changes, the projected result is often imperfect and cannot be directly used as the final editing output.

To address this, 
we treat the projection as a learnable structural prior rather than a deterministic constraint.
Specifically, instead of directly replacing image content with the projected result, we convert $\mathbf{P}_{i-1 \rightarrow i}$ into residual structural features and inject them into the diffusion transformer through a dedicated conditioning path. This allows the model to adaptively exploit cross-view structural cues while preserving the generation flexibility. The editing process is thus formulated as:

\begin{equation}
\hat{\mathbf{I}}_i = \mathcal{E}(\mathbf{I}_i, c \mid \mathbf{P}_{i-1 \rightarrow i}),
\end{equation}
where $\mathcal{E}(\cdot)$ denotes the image editor conditioned on both the text prompt and the projected structural guidance.

To incorporate the projected structural cue into the backbone model, we introduce a lightweight structural conditioning path that shares the same architecture as the diffusion backbone but with significantly fewer blocks. Let the backbone consist of $N$ DiT (diffusion transformer) blocks and the structural path consist of $M$ blocks, where $M \ll N$. 


The structural path takes as input the current view $\mathbf{I}_i$, the text prompt $c$, and the projected cue $\mathbf{P}_{i-1 \rightarrow i}$. Here, $\mathbf{I}_i$ and $c$ provide the same contextual space as the backbone, while the projected cue supplies the cross-view structural prior. The path then produces a sequence of intermediate structural features $\{\mathbf{v}_k\}_{k=0}^{M-1}$, each of which has the same dimensionality as the backbone hidden states.

To efficiently inject structural guidance, we adopt a block-wise residual conditioning strategy. Specifically, each structural feature $\mathbf{v}_k$ is shared across a group of backbone blocks and added to their hidden states. Formally, the hidden state at the $i$-th backbone block is updated as:
\begin{equation}
\mathbf{h}_i’ \leftarrow \mathbf{h}_i + \mathbf{v}_{\lfloor i / r \rfloor},
\end{equation}
where $r = \lceil N / M \rceil$ denotes the block interval, $\mathbf{h}_i$ represents the original output of block $i$. 

This design enables multi-level structural control with few additional parameters. By distributing conditioning signals across network depth, it incorporates geometry-aware guidance efficiently while preserving the generative capacity of the original backbone.

\vspace{-0.2in}
\subsection{Patch-Level Semantic Propagation}
\vspace{-0.05in}
\label{semantic}

While projection-guided structural guidance enforces geometric alignment across views, it does not ensure stable edited content under viewpoint changes. 
As a result, independently edited views often exhibit \emph{semantic drift}, where color, material, or local style varies across views even when geometry is roughly aligned. To address this issue, we introduce a patch-level semantic propagation path to propagate the edited semantics across views.

Specifically, we encode the edited result of the previous view $\hat{\mathbf{I}}_{i-1}$ into spatial feature tokens using a frozen encoder. These patch-level tokens preserve fine-grained local semantics and serve as a compact semantic reference for the accumulated editing state. Although only the immediately preceding view is used, it already carries semantic information propagated from earlier views through the sequential editing process, therefore provides sufficient context for the current view.

Let $\mathbf{F}$ denote the extracted reference features.
For a DiT block $l$, we project reference features into key and value representations through learnable linear transformations:
\begin{equation}
\mathbf{K}^{(l)'} = \mathbf{W}^{(l)}_k \mathbf{F}, \qquad \mathbf{V}^{(l)'} = \mathbf{W}^{(l)}_v \mathbf{F},
\end{equation}
where $\mathbf{W}^{(l)}_k$ and $\mathbf{W}^{(l)}_v$ are learnable parameters.

let $\mathbf{A}^{(l)}$ denote its original attention output computed from the current view. To realize semantic propagation, we augment the attention computation with a reference-guided term, where the query $\mathbf{Q}^{(l)}$ attends to the propagated reference features:
\begin{equation}
\mathrm{Attn}_{\mathrm{ref}}(\mathbf{Q}^{(l)}, \mathbf{K}^{(l)'}, \mathbf{V}^{(l)'}) 
= 
\mathrm{Softmax}\!\left(\frac{\mathbf{Q}^{(l)} \mathbf{K}^{(l)'\top}}{\sqrt{d}}\right)\mathbf{V}^{(l)'}.
\end{equation}
The enhanced attention output is then given by
\begin{equation}
\tilde{\mathbf{A}}^{(l)} = \mathbf{A}^{(l)} + \alpha \cdot \mathrm{Attn}_{\mathrm{ref}}(\mathbf{Q}^{(l)}, \mathbf{K}^{(l)'}, \mathbf{V}^{(l)'}) ,
\label{eq:attention}
\end{equation}
where $\alpha$ is a scaling parameter, $l \in \mathcal{S}$ and $\mathcal{S}$ denotes the selected set of intermediate backbone blocks.

We instantiate the semantic propagation path only in intermediate layers, where semantic abstraction is most suitable for cross-view propagation. Early layers mainly capture low-level signals, while very late layers are more tightly coupled with view-specific synthesis. This design stabilizes semantic propagation and improves parameter and training efficiency.

This design enables the current view to selectively retrieve and reuse previously established semantic attributes, allowing the model to maintain consistent appearance properties such as color and material across viewpoints. By operating at the feature level, the proposed mechanism propagates semantic state rather than enforcing explicit alignment, thereby complementing the projection-guided structural guidance path and effectively reducing semantic drift in multi-view editing.

\vspace{-0.2in}
\subsection{Consistency-Aware Multi-View Editing Dataset}
\vspace{-0.05in}
\label{data}

\begin{figure*}[htbp]
  \centering
  \includegraphics[width=\textwidth]{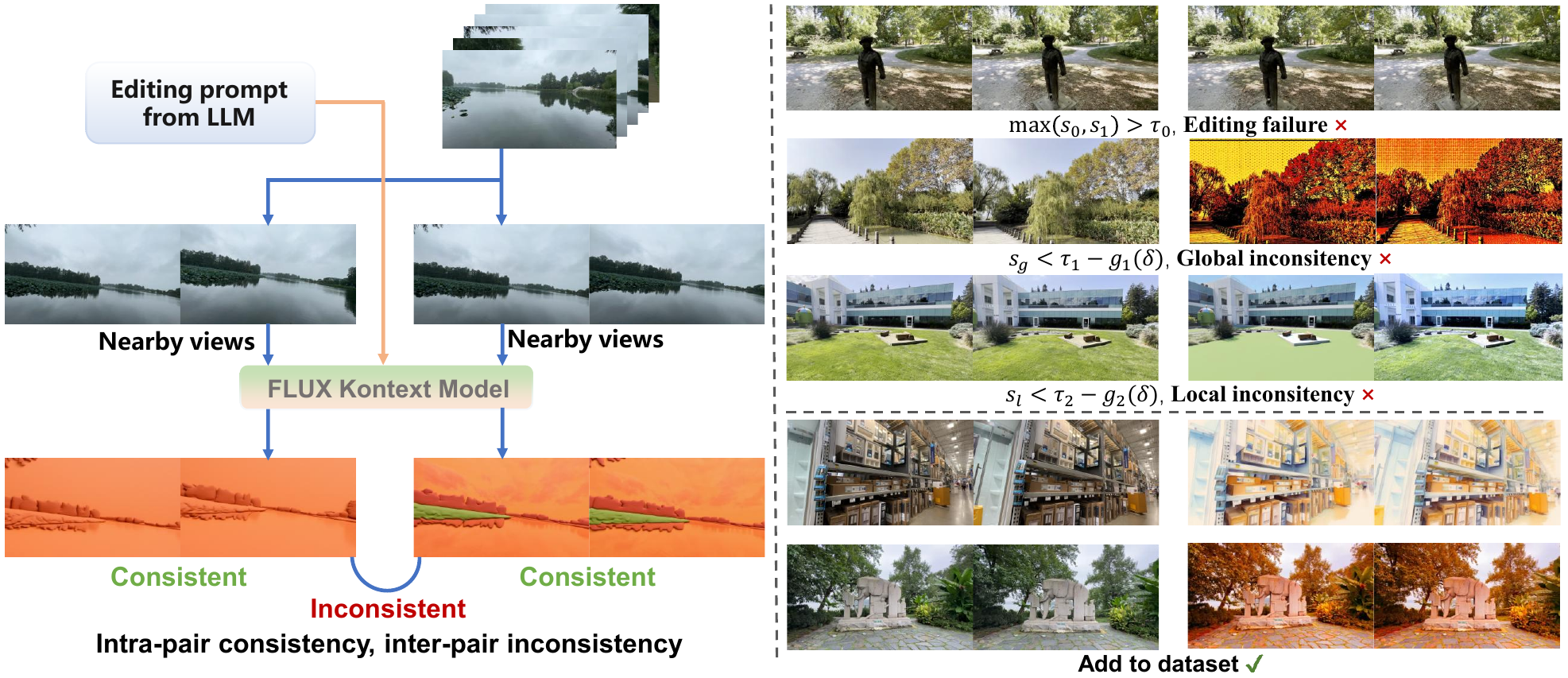}
  \vspace{-0.2in}
\caption{
Consistency-aware multi-view editing dataset construction. Left: view pairs with limited viewpoint difference are concatenated and jointly edited by a pre-trained image editing model, providing pairwise consistent supervision. Right top: failed or inconsistent edited pairs are discarded according to edit validity, global consistency, and local consistency. Right bottom: the retained pairs form the final training set for cross-view consistency learning.
}
  \label{fig:dataset}
  \vspace{-0.2in}
\end{figure*}

The proposed structural guidance and semantic propagation paths require supervision that explicitly encodes cross-view relationships. Standard image editing datasets only specify how a single image should be edited, but do not capture how edited content should remain consistent across viewpoints. We observe that a strong image editor can already produce locally consistent edits when two nearby views are concatenated and edited jointly. However, such consistency is limited to the edited pair itself and does not directly extend to an entire view sequence. 
This is because joint editing only constrains the current pair, and image editors are limited by input resolution, making it impractical to jointly process long multi-view sequences.
Motivated by this property, we construct a consistency-aware paired dataset that serves as a supervision source for learning cross-view consistency.

We build the dataset from video sequences in \cite{ling2024dl3dv, xia2024rgbd}. For each sequence, we sample a pair of frames $(\mathbf{I}_x, \mathbf{I}_y)$ with limited viewpoint difference $\delta$. Given an editing instruction $c$ generated by large language models, we concatenate the two images at the pixel level and apply the editing model:
\begin{equation}
(\hat{\mathbf{I}}_x, \hat{\mathbf{I}}_y) = \mathcal{E}(\mathrm{Concat}(\mathbf{I}_x, \mathbf{I}_y), c),
\end{equation}
where $\mathcal{E}$ denotes the image editing operator. This process provides pairwise edited results with relatively strong intra-pair consistency. 

For the retained edited pairs, we further estimate depth and camera pose using Depth Anything 3 (DA3), and compute cross-view projections in the edited domain:
\begin{equation}
\mathbf{P}_{x \rightarrow y} = \mathcal{W}(\hat{\mathbf{I}}_x, \mathbf{D}_x, \pi_x, \pi_y), 
\mathbf{P}_{y \rightarrow x} = \mathcal{W}(\hat{\mathbf{I}}_y, \mathbf{D}_y, \pi_y, \pi_x),
\end{equation}
where $\mathbf{D}_x$ and $\mathbf{D}_y$ represents the depth map estimated by DA3, $\mathcal{W}(\cdot)$ denotes a warping operation parameterized by depth and relative camera pose. 

Since jointly edited pairs may still contain editing failures or inconsistent results, we introduce a multi-criteria filtering strategy based on DINOv3 \cite{simeoni2025dinov3} features. First, to remove failed edits, we compute the similarity between each original image and its edited counterpart:
\begin{equation}
s_0 = \mathrm{sim}(\mathbf{I}_x, \hat{\mathbf{I}}_x), 
\qquad
s_1 = \mathrm{sim}(\mathbf{I}_y, \hat{\mathbf{I}}_y).
\end{equation}
We discard the pair if
\begin{equation}
\max(s_0, s_1) > \tau_0,
\end{equation}
since this indicates that at least one side remains overly similar to the original image, suggesting an editing failure and making the pair unreliable as supervision.

We then evaluate cross-view consistency in the edited domain using both global and local DINOv3 features. Here, the DINOv3 similarities are used only for sample quality filtering, rather than for exact geometric correspondence estimation. Let $f_g(\cdot)$ and $f_l(\cdot)$ denote the global and local features, respectively. The global similarity is defined as
\begin{equation}
s_g = \cos\big(f_g(\hat{\mathbf{I}}_x), f_g(\hat{\mathbf{I}}_y)\big),
\end{equation}
where $\cos$ indicates the cosine similarity. For local consistency, we compute bidirectional patch matching:
\begin{equation}
\begin{aligned}
s_l = \frac{1}{2N} \Bigg(
&  \sum_{p=1}^N \max_q \langle f_l^p(\hat{\mathbf{I}}_x), f_l^q(\hat{\mathbf{I}}_y) \rangle \\
+\, &  \sum_{q=1}^N \max_p \langle f_l^q(\hat{\mathbf{I}}_y), f_l^p(\hat{\mathbf{I}}_x) \rangle
\Bigg),
\end{aligned}
\end{equation}
where $N$ denotes the number of patch tokens, and $f_l^p(\cdot)$ denotes the $\ell_2$-normalized feature of the $p$-th token.

To account for viewpoint variation, we adopt adaptive thresholds conditioned on camera viewpoint difference $\delta$:
\begin{equation}
s_g > \tau_1 - g_1(\delta), 
\qquad
s_l > \tau_2 - g_2(\delta),
\end{equation}
where $g_1$ and $g_2$ are monotonically increasing functions. Samples that violate these conditions are discarded.

The retained samples constitute the final training set:
\begin{equation}
\mathcal{T} = (\mathbf{I}_x, \mathbf{I}_y, \hat{\mathbf{I}}_x, \hat{\mathbf{I}}_y, \mathbf{P}_{x \rightarrow y}, \mathbf{P}_{y \rightarrow x}, c).
\end{equation}
Each training tuple provides the original neighboring pair, the jointly edited pair, and the edited-domain cross-view projections, thereby supplying supervision for both the semantic propagation path and the structural guidance path. Fig.~\ref{fig:dataset} illustrates the overall dataset construction process.

\vspace{-0.2in}
\subsection{Training \& 3DGS Update}
\vspace{-0.05in}
\label{training}

\begin{figure*}[htbp]
  \centering
  \includegraphics[width=\textwidth]{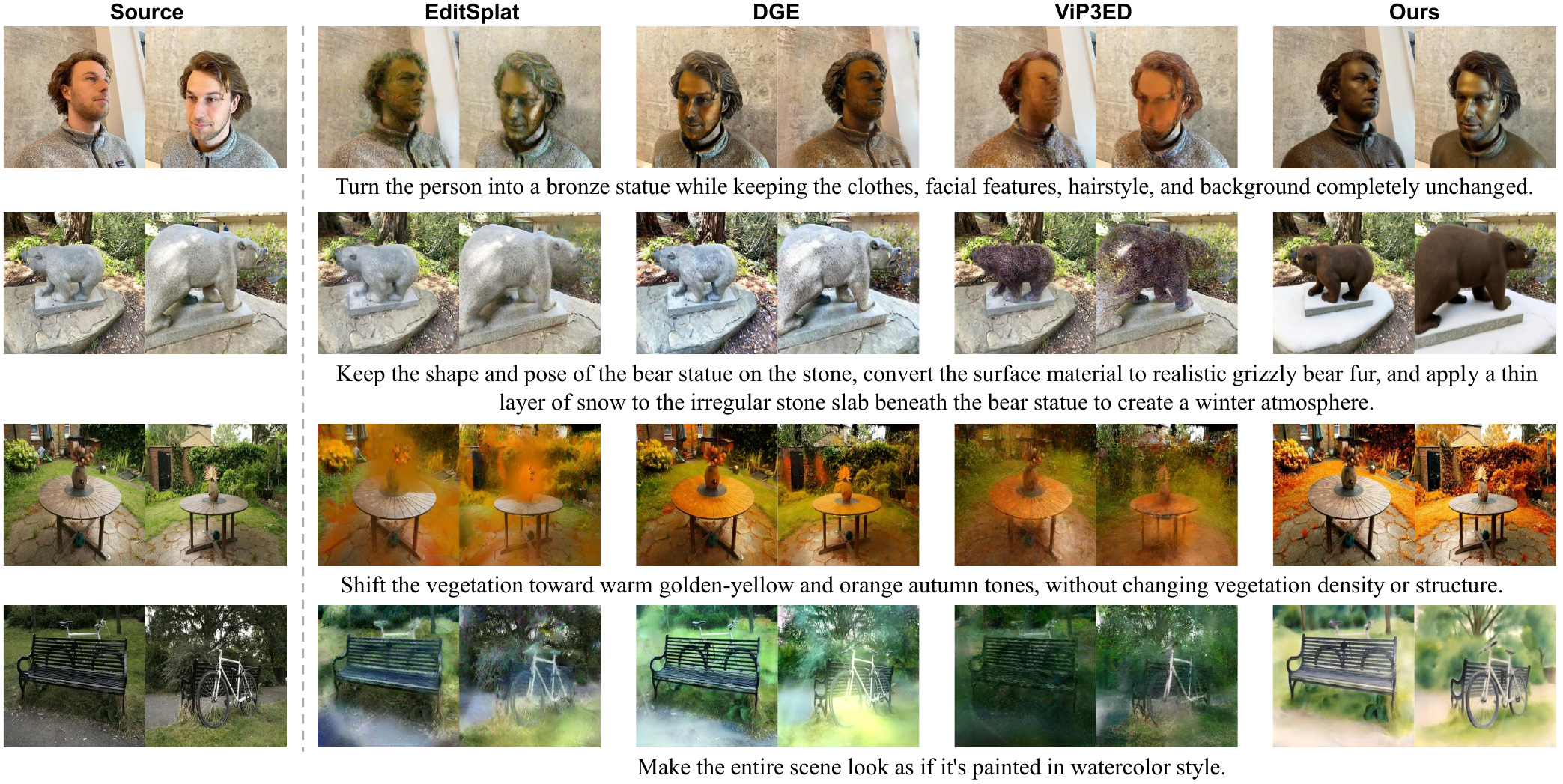}
  \vspace{-0.2in}
\caption{
Qualitative comparison with state-of-the-art methods under various editing prompts. 
The leftmost column shows source images, while the right columns show rendering images from edited 3DGS.
Our method achieves more consistent, coherent, and faithful editing results across viewpoints, particularly for complex scene-level editing instructions.
}
  \label{fig:compare}
  \vspace{-0.2in}
\end{figure*}

\subsubsection{Training}

We adopt the latent flow-matching formulation used in \cite{labs2025flux}. Let $\mathbf{z}_0 \sim \mathcal{N}(0, I)$ denote Gaussian noise, and let $\mathbf{z}_1$ denote the latent representation of the target edited image encoded by a pretrained VAE. Following the standard linear interpolation path, the intermediate latent state is defined as
\begin{equation}
\mathbf{z}_t = (1-t)\mathbf{z}_0 + t\mathbf{z}_1,
\label{eq:zt}
\end{equation}
where $t\sim \mathcal{U}(0,1)$, The model is trained to predict the target velocity field along this trajectory, and for consistency with our notation, the supervision target is written as $\mathbf{z}_0-\mathbf{z}_1$.

We train the proposed framework in two stages under a shared overall architecture with different trainable parameter subsets. We adopt this stage-wise strategy to decouple geometry-aware structural learning from semantic propagation. We first learn structural control under explicit cross-view correspondence, and then freeze it to provide stable structural guidance while training the semantic propagation path. In this way, the two pathways are learned progressively instead of being jointly optimized from the beginning, reducing interference between structural alignment and semantic continuity.

In the first stage, the trainable parameters belong to the projection-guided structural guidance, denoted by $\theta_s$, while the  semantic propagation path remains inactive. Given the current image $\mathbf{I}_y$, the text instruction $c$, and the projected structural cue $\mathbf{P}_{x\rightarrow y}$ obtained by warping the edited reference view $\hat{\mathbf{I}}_x$ into the current viewpoint, the structural guidance path converts the projected cross-view cue into residual structural conditioning signals and injects them into the frozen backbone. The resulting velocity prediction is denoted by:
\begin{equation}
v_{\theta_s}(\mathbf{z}_t, \mathbf{I}_y, \mathbf{P}_{x\rightarrow y}, c).
\end{equation}

The corresponding flow-matching objective is:
\begin{equation}
\mathcal{L}_{\mathrm{s}}
=
\left\|
v_{\theta_s}(\mathbf{z}_t, \mathbf{I}_y, \mathbf{P}_{x\rightarrow y}, c) - (\mathbf{z}_0-\mathbf{z}_1)
\right\|_2^2,
\end{equation}
where $\mathbf{z}_t$ is computed according to Eq.~\eqref{eq:zt}, and $\mathbf{z}_1$ is the latent code of the target edited image $\hat{\mathbf{I}}_y$. This stage enables the model to learn reliable projection-guided structural control under explicit cross-view correspondence.

In the second stage, we freeze the structural guidance path and train the semantic propagation path, whose trainable parameters are denoted by $\theta_m$. The current image $\mathbf{I}_y$, the text instruction $c$, and the projected cue $\mathbf{P}_{x\rightarrow y}$ are still retained, while the edited reference image $\hat{\mathbf{I}}_x$ is additionally encoded into patch-level features. These features are injected into selected transformer layers through trainable K/V linear layers.
The predicted velocity in this stage is written as:
\begin{equation}
v_{\theta_m}(\mathbf{z}_t, \mathbf{I}_y, \hat{\mathbf{I}}_x, \mathbf{P}_{x\rightarrow y}, c).
\end{equation}

The Stage-2 objective is defined as:
\begin{equation}
\mathcal{L}_{\mathrm{m}}
=
\left\|
v_{\theta_m}(\mathbf{z}_t, \mathbf{I}_y, \hat{\mathbf{I}}_x, \mathbf{P}_{x\rightarrow y}, c) - (\mathbf{z}_0-\mathbf{z}_1)
\right\|_2^2.
\end{equation}

Compared with Stage 1, Stage 2 freezes the structural guidance path and trains the semantic propagation path to improve cross-view semantic consistency.

\subsubsection{3D Gaussian Splatting Update}

At inference time, we perform sequential multi-view editing to propagate both projection-guided structural guidance and patch-level semantic propagation. Given a set of input views $\{\mathbf{I}_t\}$ rendered from the 3D representation with known camera poses, the editing process follows a predefined order over views.

For the first view, the edited result $\hat{\mathbf{I}}_1$ is directly generated by the pretrained backbone conditioned on the input image and text instruction. 
For subsequent views, the editing is performed recursively. Specifically, for the current view $t$, the projected cue $\mathbf{P}_{t-1 \rightarrow t}$ derived from the previous edited view is fed into the structural guidance path. 
Meanwhile, the previous edited image $\hat{\mathbf{I}}_{t-1}$
is encoded into patch-level semantic features, which are propagated to the current view through the semantic propagation path.
The edited result of the current view is then written as:
\begin{equation}
\hat{\mathbf{I}}_t = \mathcal{E'}(\mathbf{I}_t, c \mid \mathbf{P}_{t-1 \rightarrow t}, \hat{\mathbf{I}}_{t-1}),
\end{equation}
where $\mathcal{E'}(\cdot)$ denotes the full editor with the trained structural guidance path and semantic propagation path.

After all views have been edited, the resulting edited images $\{\hat{\mathbf{I}}_t\}$ are used to optimize the 3D Gaussian Splatting representation according to Eq.~\eqref{eq:optimizegs}. Fig.~\ref{fig:framework2} illustrates the overall training and inference pipeline.

%% file: 4_experiments.tex
\begin{table}[t]
\centering
\normalsize
\caption{Quantitative comparison with other 3D editing methods. CLIP$_{\text{sim}}$: CLIP text-image similarity; CLIP$_{\text{dir}}$: CLIP directional similarity.}
\vspace{-0.05in}
\label{tab:quantitative}
\begin{tabular}{ccccc}
\hline
Method & Year & CLIP$_{\text{sim}}$ $\uparrow$ & CLIP$_{\text{dir}}$ $\uparrow$ & DINO $\uparrow$ \\
\hline
GaussCtrl \cite{wu2024gaussctrl} & 2024 & 0.2287 & 0.0478 & 0.7278 \\
DGE \cite{chen2024dge}      & 2024 & 0.2384 & 0.0656 & 0.7501 \\
EditSplat \cite{lee2025editsplat} & 2025 & 0.2366 & 0.0612 & 0.7353 \\
ViP3DE \cite{chen2026fast}  & 2026 & 0.2278 & 0.0588 & 0.7488 \\
\hline
Ours      & 2026 & \textbf{0.2561} & \textbf{0.0874} & \textbf{0.7768} \\
\hline
\end{tabular}
\vspace{-0.2in}
\end{table}

\vspace{-0.05in}
We conduct comprehensive experiments to evaluate both editing quality and multi-view consistency. 
We first describe the experimental setup, including evaluation dataset, implementation details, evaluation metrics, and baseline methods in Sec. \ref{setup}. We then compare our method with the baseline methods through quantitative and qualitative results in Sec. \ref{compare}. Finally, we perform ablation studies to analyze the effectiveness of each path and demonstrate the necessity of our consistency-aware design in Sec. \ref{ablation}.

\vspace{-0.2in}
\subsection{Experimental Setup}
\vspace{-0.05in}
\label{setup}
\subsubsection{Evaluation Dataset}

To assess the 3D scene editing capability of our method, we curate evaluation scenes from the IN2N \cite{haque2023instruct}, Mip-NeRF 360 \cite{barron2022mip}, BlendedMVS \cite{yao2020blendedmvs}, and LLFF \cite{mildenhall2019local} datasets. 
For each scene, editing is performed using the same text prompt, after which the edited scene is rendered from uniformly sampled camera poses. 

\subsubsection{Implementation Details}

Our method is built upon the FLUX Kontext backbone \cite{labs2025flux}, which consists of 57 DiT blocks, including 19 double blocks and 38 single blocks. All experiments are conducted at a resolution of $1024 \times 1024$, following the predefined resolution settings of FLUX Kontext.

We adopt a two-stage training strategy, as mentioned in Sec. \ref{training}, while keeping the original backbone frozen throughout both stages. In the first stage, we train the structural guidance path composed of 3 double blocks and 6 single blocks, initialized from the first 3 double blocks and the first 6 single blocks of the backbone. A zero-initialized linear layer is appended after each block output. 
In the second stage, the structural guidance path stays frozen, active only in forward propagation.
We propagate reference features into the attention layers from the 7th to 40th DiT blocks by introducing two additional learnable linear layers. The attention weight coefficient $\alpha$ in Eq.~(\ref{eq:attention}) is set to 0.4. During inference, both paths are enabled.

During dataset construction, we use thresholds $\tau_0=0.97$, $\tau_1=0.93$, and $\tau_2=0.9$ for sample filtering. We only consider view pairs with viewpoint difference smaller than 0.75. The angle-dependent thresholds are further defined as:
\[
g_1(\theta)=((\theta/0.75)^{0.67})\times 0.08,
g_2(\theta)=((\theta/0.75)^{0.67})\times 0.12.
\]
While this filtering strategy may occasionally exclude some good samples, it substantially reduces problematic cases in the dataset, leading to cleaner supervision and more reliable training. In total, our dataset comprises 280K samples.

For optimization, we use AdamW with a learning rate of $2\times10^{-5}$ in the first stage and $1.6\times10^{-5}$ in the second stage. Training is performed with a total batch size of 32 on 8 NVIDIA H20 GPUs with bfloat16 precision and gradient checkpointing for memory efficiency. The first stage runs for 22k iterations, taking approximately 6 days, while the second stage runs for 24k iterations and takes about 6.5 days. Inference is conducted on a single NVIDIA H20 GPU.

\subsubsection{Evaluation Metrics}

We quantitatively evaluate the results using three metrics, 
CLIP text-image similarity \cite{brooks2023instructpix2pix} represents the cosine similarity between the text and image embeddings encoded by CLIP, 
CLIP directional similarity \cite{gal2022stylegan} represents the cosine similarity between the image and text editing directions,
and DINO similarity \cite{simeoni2025dinov3} represents the cosine similarity between edited renderings for measuring cross-view consistency.

\subsubsection{Baseline Methods}

We compare our method with recent 3DGS-based state-of-the-art baselines, including GaussCtrl \cite{wu2024gaussctrl}, DGE \cite{chen2024dge}, EditSplat \cite{lee2025editsplat}, and ViP3DE \cite{chen2026fast}. 
GaussCtrl improves consistency through depth-conditioned editing and latent alignment; EditSplat uses multi-view fusion guidance; DGE leverages spatio-temporal attention with geometric constraints; and ViP3DE exploits video priors for inter-frame consistency. 
These methods represent strong existing approaches for view-consistent 3D editing.

\vspace{-0.15in}
\subsection{Quantitative \& Qualitative Comparison}
\label{compare}

\begin{table}[t]
\centering
\normalsize
\caption{Ablation study on the key components of our method. The full model achieves the best overall performance, showing that both structural guidance and semantic propagation contribute to consistent multi-view editing.}
\vspace{-0.05in}
\label{tab:ab_module}
\begin{tabular}{cccc}
\hline
Method & CLIP$_{\text{sim}}$ $\uparrow$ & CLIP$_{\text{dir}}$ $\uparrow$ & DINO $\uparrow$ \\
\hline
Direct Edit & 0.2269 & 0.0463 & 0.7423 \\
w/o struct. transfer & 0.2403 & 0.0733 & 0.7592 \\
w/o sem. mem. inject & 0.2442 & 0.0742 & 0.7633 \\
\hline
Full Model      & \textbf{0.2561} & \textbf{0.0874} & \textbf{0.7768} \\
\hline
\end{tabular}
\vspace{-0.2in}
\end{table}

\begin{figure}[t]
  \centering
  \includegraphics[width=0.48\textwidth]{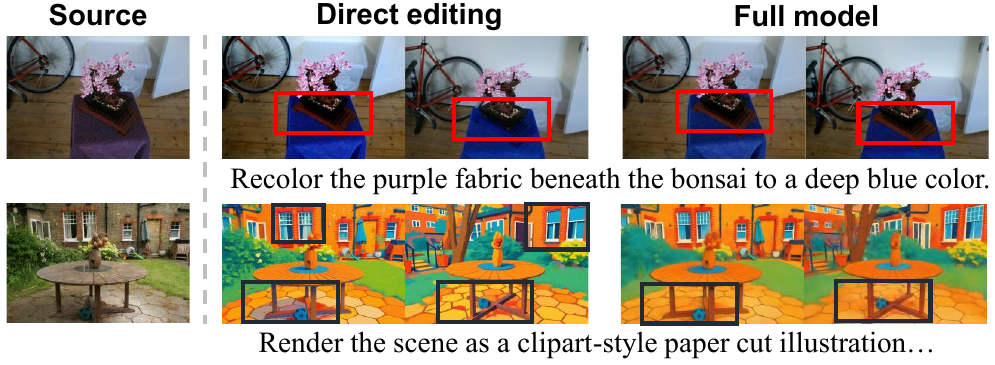}
  \vspace{-0.05in}
  \caption{Visual comparison between Direct Editing and our full model. Direct Editing performs per-view editing independently, causing appearance inconsistency across viewpoints, while our full model produces more coherent results.}
  \label{fig:ab1}
  \vspace{-0.1in}
\end{figure}

To evaluate the effectiveness of our method, we provide both quantitative and qualitative comparisons with the baseline methods in Table~\ref{tab:quantitative} and Fig. \ref{fig:compare}, respectively.

\subsubsection{Quantitative Comparison} 

As shown in Table~\ref{tab:quantitative}, our method outperforms all baselines across all metrics. Specifically, our approach achieves the highest CLIP similarity, indicating better alignment between the edited images and the target text prompts. Moreover, our method obtains a significantly higher CLIP directional similarity, demonstrating that the editing direction is more consistent with the semantic transformation specified by the text. In terms of DINO similarity, our method achieves the best performance, reflecting stronger cross-view consistency compared to existing approaches.

\subsubsection{Qualitative Comparison} 

As illustrated in Fig. \ref{fig:compare}, existing methods often suffer from inconsistent appearance and structural distortions across different viewpoints. For example, methods such as DGE and ViP3DE may produce unstable geometry or inconsistent textures under viewpoint changes, while EditSplat tends to introduce over-smoothed or blurred results. In contrast, our method preserves both structural integrity and semantic consistency across views. For instance, in the bear statue example, our method successfully maintains consistent fur appearance and geometric structure across viewpoints, while other methods exhibit noticeable inconsistencies or artifacts. Similarly, for style-based edits such as watercolor or pencil sketch transformations, our approach produces more coherent and visually stable results across multiple views.

Overall, both quantitative and qualitative results demonstrate that our method achieves superior performance in maintaining multi-view consistency while preserving high-quality semantic editing. 
Notably, under complex scene-level editing instructions, our method produces more complete and refined edits that better fulfill the target requirements, while maintaining cross-view consistency and avoiding noticeable noise and drift.

\vspace{-0.2in}
\subsection{Ablation Study}
\vspace{-0.05in}
\label{ablation}
\subsubsection{Effect of Key Components} 

\begin{figure}[t]
  \centering
  \includegraphics[width=0.42\textwidth]{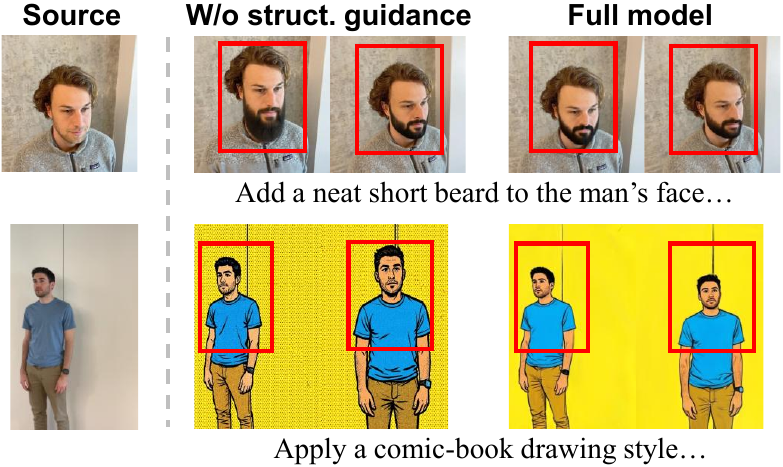}
  \vspace{-0.05in}
  \caption{Visual comparison between w/o structural guidance and the full model. Removing the structural guidance path weakens structural stability across viewpoints, leading to less reliable geometric correspondence and local misalignment.}
  \label{fig:ab2}
  \vspace{-0.2in}
\end{figure}

\begin{figure}[t]
  \centering
  \includegraphics[width=0.48\textwidth]{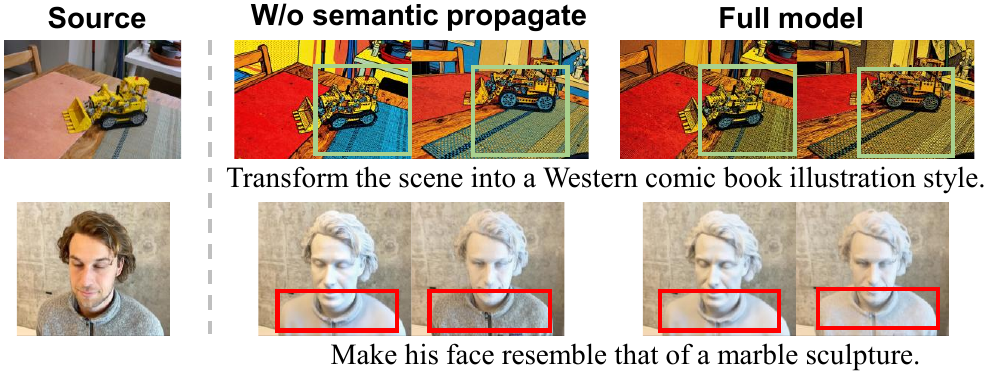}
  \vspace{-0.05in}
  \caption{Visual comparison between w/o semantic propagation and the full model. W/o semantic propagation, the  edits become less complete and less consistently across viewpoints, whereas the full model yields more coherent semantic changes.}
  \label{fig:ab3}
  \vspace{-0.2in}
\end{figure}

\begin{table}[t]
\centering
\normalsize
\caption{Analysis of the structural guidance path design. D: double blocks, S: single blocks.}
\vspace{-0.05in}
\label{tab:ab_struct_design}
\begin{tabular}{ccccc}
\hline
Setting & CLIP$_{\text{sim}}$ $\uparrow$ & CLIP$_{\text{dir}}$ $\uparrow$ & DINO $\uparrow$ & \makecell{Params} \\
\hline
2D + 4S & 0.2489 & 0.0789 & 0.7638 & 1.33B \\
3D + 6S & \textbf{0.2561} & 0.0874 & \textbf{0.7768} & 1.99B \\
4D + 8S & 0.2558 & \textbf{0.0880} & 0.7752 & 2.66B \\
\hline
\end{tabular}
\vspace{-0.25in}
\end{table}

We conduct ablation studies on the key components of our framework, including Direct editing, removing the structural guidance path, and removing the semantic propagation path, while using the full model as the reference. The overall quantitative results are summarized in Table~\ref{tab:ab_module}.
We then analyze each ablation setting together with its corresponding visual comparison.

The setting of direct editing, corresponding to the first row of Table~\ref{tab:ab_module}, edits each view independently without any consistency modeling. As shown in Fig. \ref{fig:ab1}, this setting leads to clear cross-view inconsistency. 
In the bonsai example, Direct Editing yields inconsistent recoloring across views and also affects nearby regions. In the clipart-style scene, the stylized results vary across viewpoints, with inconsistent colors.
In contrast, the full model produces much more coherent results across views. These observations show that  per-view editing is insufficient for multi-view consistent scene editing.

The variant without the structural guidance path corresponds to the second row of Table~\ref{tab:ab_module}. As shown in Fig. \ref{fig:ab2}, removing this path weakens structural stability across viewpoints. For example, in the beard editing case, the beard shape and placement become less consistent across views. A similar issue can also be observed in the comic-style example, where local contours and structural details are less stable than those of the full model. This indicates that the structural guidance path is important for preserving reliable geometric correspondence across viewpoints.

The variant without semantic propagation path is reported in the third row of Table~\ref{tab:ab_module}. As shown in Fig. \ref{fig:ab3}, removing this path affects the completeness and consistency of semantic editing. In the western comic style example, the target stylization is less consistently propagated across viewpoints. In the marble sculpture example, the transformation of the clothes material is also weaker and less stable than that of the full model. These results suggest that semantic memory injection is essential for propagating target semantic attributes and ensuring more complete edits across views.

\vspace{-0.05in}
\subsubsection{Design Choices}

\begin{table}[t]
\centering
\normalsize
\caption{Analysis of the semantic propagation location.}
\vspace{-0.05in}
\label{tab:ab_sem_design}
\begin{tabular}{cccc}
\hline
Location  & CLIP$_{\text{sim}}$ $\uparrow$ & CLIP$_{\text{dir}}$ $\uparrow$ & DINO $\uparrow$ \\
\hline
1-34 & 0.2501 & 0.0823 & 0.7723 \\
7-40  & \textbf{0.2561} & \textbf{0.0874} & \textbf{0.7768} \\
24-57    & 0.2513 & 0.0843 & 0.7699 \\
\hline
\end{tabular}
\vspace{-0.2in}
\end{table}

We further analyze two key design choices: the architecture of the structural guidance path and the injection location of the semantic propagation path.

\textbf{Structural guidance architecture.} Table~\ref{tab:ab_struct_design} compares different path depths. The shallow design (2D + 4S) has limited capacity for cross-view structural correspondence, while the deeper design (4D + 8S) brings only marginal gains with more parameters. The 3D + 6S design improves all metrics and offers a better balance between effectiveness and complexity.

\textbf{Semantic propagation location.} Table~\ref{tab:ab_sem_design} compares shallow (1--34), middle (7--40), and deep (24--57) injection ranges. Shallow layers capture low-level patterns, whereas deep layers introduce semantic guidance too late. Middle-layer injection achieves the best results, indicating a suitable balance between semantic propagation and stable multi-view editing.

%% file: 5_conclusion.tex
\vspace{-0.05in}
In this paper, we proposed a cross-view consistency framework for 3D scene editing. Instead of treating multi-view editing as independent single-image editing, we reformulated it as a cross-view dependent process that approximates the joint distribution required for view-consistent scene editing. This formulation substantially alleviates multi-view inconsistency and enables more complex scene-level modifications.
To realize this framework, we introduced a dual-path consistency mechanism that separately models projection-guided structural guidance and patch-level semantic propagation, together with a consistency-aware paired multi-view editing dataset, CVC-Edit, for reliable supervision. Extensive experiments demonstrated that our method achieves superior editing quality and cross-view consistency compared with existing approaches. 